# Exploring the Vulnerabilities of Machine Learning and Quantum Machine Learning to Adversarial Attacks using a Malware Dataset: A Comparative Analysis

Mst Shapna Akter*, Hossain Shahriar[†], Iysa Iqbal[‡], MD Hossain[§], M.A. Karim[¶], Victor Clincy[∥], Razvan Voicu[**]

*Department of Computer Science, Kennesaw State University, USA
Email: makter2@students.kennesaw.edu

[†]Department Information Technology, Kennesaw State University, USA
Email: hshahria@kennesaw.edu

[‡]Nicolet High School, USA
Email: iysa.iqbal@gmail.com

[§]Southern Polytechnic College of Engineering and Engineering Technology, USA
Email: mhossa14@kennesaw.edu

[¶]Southern Polytechnic College of Engineering and Engineering Technology, USA
Email: mkarim4@kennesaw.edu

[∥]Department of Computer Science, Kennesaw State University, USA
Email: Vclincy@kennesaw.edu

[**]Department of Engineering, Kennesaw State University, USA
Email: rvoicu@kennesaw.edu

*Abstract*—The burgeoning fields of machine learning (ML) and quantum machine learning (QML) have shown remarkable potential in tackling complex problems across various domains. However, their susceptibility to adversarial attacks raises concerns when deploying these systems in security-sensitive applications. In this study, we present a comparative analysis of the vulnerability of ML and QML models, specifically conventional neural networks (NN) and quantum neural networks (QNN), to adversarial attacks using a malware dataset. We utilize a software supply chain attack dataset known as ClaMP and develop two distinct models for QNN and NN, employing Pennylane for quantum implementations and TensorFlow and Keras for traditional implementations. Our methodology involves crafting adversarial samples by introducing random noise to a small portion of the dataset and evaluating the impact on the models' performance using accuracy, precision, recall, and F1 score metrics. Based on our observations, both ML and QML models exhibit vulnerability to adversarial attacks. While the QNN's accuracy decreases more significantly compared to the NN after the attack, it demonstrates better performance in terms of precision and recall, indicating higher resilience in detecting true positives under adversarial conditions. We also find that adversarial samples crafted for one model type can impair the performance of the other, highlighting the need for robust defense mechanisms. Our study serves as a foundation for future research focused on enhancing the security and resilience of ML and QML models, particularly QNN, given its recent advancements. A more extensive range of experiments will be conducted to better understand the performance and robustness of both models in the face of adversarial attacks.

## I. INTRODUCTION

The growing prevalence of software supply chain security threats has driven researchers to explore innovative approaches to detect and predict vulnerabilities and suspicious behaviors [1]. Software Supply Chain (SSC) attacks occur when cyber threat actors penetrate a vendor's network and insert malicious code that compromises the software before distribution to customers. Such attacks can have severe consequences for software users across sectors by gaining control over the software's regular functionality [2]. Machine Learning (ML) has long been employed as a powerful tool to address these challenges, but the exponential growth of data worldwide necessitates alternative solutions for proactive prevention and early detection of security threats. Quantum Machine Learning (QML), which leverages quantum computing concepts and quantum random access memory (QRAM), has emerged as a promising solution to handle large-scale data processing [3]. These unique characteristics have led to the increasing adoption of quantum computing in various technological fields. However, the vulnerability of ML and QML models to adversarial attacks raises concerns when deploying these systems in security-sensitive applications. Adversarial attacks in machine learning (ML) and quantum machine learning (QML) are malicious attempts to exploit the vulnerabilities of ML and QML models by generating specially crafted input samples, known as adversarial examples. These examples are designed to be imperceptibly different from the original data but can cause the models to make incorrect predictions or classifications with high confidence [4]. Adversarial attacks can be classified based on their intended goals, such as targeted attacks, which aim to manipulate the model into assigning a specific incorrect label, and untargeted attacks, which aim to cause any misclassification. Adversarial attacks in QML leverage the unique properties of quantum computing, such as superposition and entanglement, to manipulate the decision-making processes of quantum machine learning models. While research on adversarial attacks in QML is still in its early stages, some studies have demonstrated the existence of adversarial examples in quantum settings and their potential to transfer between classical and quantum models. These attacks pose a significant challenge for the deployment of ML and QML models in security-sensitive applications, such as autonomous vehicles, facial recognition systems, and cybersecurity. Consequently, adversarial attacks undermine the reliability and security of systems powered by ML and QML in critical applications, such as healthcare, finance, and cybersecurity [5]. In recent years, adversarial attacks have led to the spread of misinformation, bypassing facial recognition systems, and even causing autonomous vehicles to misinterpret road signs [6]. As a result, there is a growing interest in developing robust defense mechanisms to mitigate the impact of adversarial attacks and ensure the reliability and security of ML and QML systems [7]. In this study, we conduct a comparative analysis of the susceptibility of ML and QML models, specifically conventional neural networks (NN) and quantum neural networks (QNN), to adversarial attacks using a malware dataset. To the best of our knowledge, this is one of the few studies focusing on the software supply chain vulnerabilities dataset using quantum machine learning. Our research utilizes Pennylane, a quantum computing platform that enables quantum differentiable programming and offers seamless integration with other QML tools, such as IBM Quantum, NumPy, and TensorFlow Quantum. The primary contributions of this research are as follows:

[1] We adopt both quantum machine learning and conventional machine learning to conduct experiments on a software supply chain attack dataset.

[2] We assess the performance of NN and QNN models under adversarial attack scenarios, comparing their vulnerability and robustness.

The rest of the paper is organized as follows: Section II presents a brief overview of related studies on quantum machine learning and traditional machine learning. Section III explains the methodology adopted for our comparative research. Section IV describes the experimental setting and results, including dataset specification and processing. Section V discusses the findings of this paper, focusing on the comparative performance of NN and QNN models under adversarial attacks. Finally, Section VI concludes the paper

## II. BACKGROUND AND LITERATURE REVIEW

In recent years, machine learning (ML) and quantum machine learning (QML) have gained significant attention in various domains, including security and malware analysis [5, 8]. However, the security of ML and QML models themselves has become a critical concern due to their vulnerability to adversarial attacks [9, 10]. This literature review aims to provide an overview of the current state of research in adversarial machine learning and quantum machine learning attacks, focusing on their application to malware datasets and the software supply chain. We also highlight the existing gaps in the literature. Adversarial machine learning attacks have been extensively studied in the literature, with a primary focus on deep learning models. Szegedy et al. [11] demonstrated that deep neural networks can be easily fooled by adding imperceptible perturbations to input data. Goodfellow et al. [4] proposed the fast gradient sign method (FGSM) for generating adversarial examples, which has become a cornerstone in the field. More recently, researchers have examined the vulnerabilities of ML models used for malware analysis. Grosse et al. [12] explored the susceptibility of neural networks to adversarial examples in malware classification and demonstrated that even minor modifications to malware samples can lead to misclassification. Demontis et al. [13] proposed the concept of adversarial attacks on graph-based machine learning models for malware detection, exposing potential vulnerabilities in these models. Finlayson et al. [14] studied the susceptibility of medical machine learning models to adversarial attacks. They demonstrated that these models are vulnerable to attacks that can cause misclassification, and highlighted the potential risks of using such models in clinical settings. The authors also proposed several methods for defending against adversarial attacks in medical machine learning. Aloraini et al. [15] investigated the threat of adversarial attacks in the context of Internet of Things (IoT) devices. They analyzed the security risks posed by adversarial machine learning attacks from an insider's perspective, highlighting the potential harm that can be caused by such attacks in critical IoT applications. The authors proposed several strategies for detecting and preventing adversarial attacks in IoT systems. Alsmadi et al. [16] conducted a literature survey on adversarial machine learning in text processing. They reviewed the current state of the field and highlighted the main challenges and opportunities in this area. The authors also discussed several approaches for defending against adversarial attacks in text processing, including deep learning-based methods, rule-based methods, and hybrid methods. Mumcu et al. [17] investigated the susceptibility of video anomaly detection systems to adversarial attacks. They demonstrated that these systems are vulnerable to attacks that can cause false alarms or stealthy attacks that can go undetected. The authors proposed several defense mechanisms for improving the robustness of video anomaly detection systems against adversarial attacks, including adversarial training and defensive distillation.

Quantum machine learning (QML) is an emerging field that leverages the computational power of quantum computers to solve complex machine learning problems [8]. While QML is still in its infancy, researchers have begun to investigate the potential vulnerabilities of QML models to adversarial attacks. West et al. [18] proposed a benchmarking framework for evaluating the adversarial robustness of quantum machine learning models at scale. The authors introduced a novel approach to generating adversarial examples for quantum machine learning models based on the qubit gradient sign method. They demonstrated the effectiveness of their benchmarking framework by evaluating the adversarial robustness of several quantum machine learning models on a variety of datasets. The authors also compared the performance of quantum machine learning models with classical machine learning models and discussed the potential applications of adversarially robust quantum machine learning models in various domains, such as finance, chemistry, and cryptography. The proposed benchmarking framework provides a valuable tool for evaluating the robustness of quantum machine learning models and can help to improve the security and reliability of quantum machine learning systems. Suryotrisongko et al. [19] investigated the adversarial robustness of a hybrid quantum-classical deep learning model for detecting domain generation algorithms (DGA) used in botnet attacks. The authors proposed a quantum-classical neural network architecture that combines a classical deep neural network with a quantum neural network, which is used to encode the input data as quantum states. They evaluated the adversarial robustness of their model by generating adversarial examples using several attack methods, including the fast gradient sign method and the Carlini-Wagner attack. The authors demonstrated that their hybrid model is more robust to adversarial attacks compared to the classical deep learning model, and they also proposed a training method based on adversarial training to improve the robustness of their model. The results suggest that hybrid quantum-classical deep learning models can provide a promising approach for detecting botnet attacks, especially in the presence of adversarial attacks. The proposed framework can be extended to other applications that require robust and secure machine learning models. Gong and Deng [20] studied the universal adversarial examples and perturbations for quantum classifiers, which are quantum machine learning models used for classification tasks. The authors introduced a method to generate universal adversarial perturbations that can be applied to multiple quantum classifiers, rather than just a single classifier. They showed that these universal perturbations can be used to construct universal adversarial examples that can fool multiple quantum classifiers with high success rates. The authors also proposed a defense mechanism based on regularization to mitigate the impact of adversarial attacks. They demonstrated the effectiveness of their approach using several quantum classification tasks, including the classification of handwritten digits and the classification of images from the MNIST dataset. The results suggest that universal adversarial examples and perturbations can pose a significant threat to the security and reliability of quantum classifiers, and that defense mechanisms based on regularization can help to improve their robustness. The proposed framework provides a valuable tool for evaluating the security of quantum machine learning models and can help to improve the design of robust and secure quantum classifiers. In the context of malware analysis, several studies have explored the use of ML and QML models to detect and classify malware. Recent works have applied deep learning techniques [21] and graph-based models [22] for malware detection. Meanwhile, QML models have been applied to various security-related tasks, such as intrusion detection and post-quantum cryptography [23]. Although adversarial machine learning attacks on malware analysis models have been studied, there is limited research on comparing the vulnerabilities of ML and QML models in this context. Furthermore, the application of adversarial attacks to QML models for malware analysis remains largely unexplored, leaving ample room for research in this area. Additionally, the software supply chain is another essential aspect to consider in the context of ML and QML vulnerabilities. While there have been some efforts to address software supply chain security using ML, the potential impact of adversarial attacks on these models remains unaddressed. Similarly, the exploration of QML-based solutions for securing the software supply chain is still in its early stages, and their potential vulnerabilities to adversarial attacks have yet to be investigated.

## III. METHODOLOGY

In this paper, we utilized Quantum Neural Network (QNN), a subfield of Quantum Machine Learning (QML), to analyze the ClaMP dataset. To ensure that our adversarial attacks are effective, we optimized each part of the dataset created from the ClaMP and introduced perturbed data during fine-tuning. We used Python and Scikit-Learn (sklearn) library functions such as shuffle, index-reset, and drop functions for data preprocessing. After applying shuffle functions, we organized the dataset in ascending order by resetting the index. We converted categorical values into numerical values and normalized all numerical values to maintain a similar scale, as quantum machine learning models require numerical values. The entire dataset comprising 5,210 rows was split into two portions: 80% for training and 20% for fine-tuning the model. For each step, we divided the data into three parts: 60% for training, 20% for validation, and 20% for testing. We added perturbed data to the 20% dataset for fine-tuning the model. The quantum machine learning model was applied to each dataset's separated portions. The features were encoded into quantum states before feeding into the QML model. The QNN framework, which originates from neurocomputing theory, combines machine learning, quantum computing, and artificial neural network concepts. It can be applied for processing neural computing using vast levels of datasets to obtain the expected result. Input data is encoded into a suitable qubit state with a proper number of qubits before being processed through QNN. The qubit state is then modified for a specific number of layers using parameterized rotation and entangling gates, with the predicted value of a Hamilton operator guiding the modified qubit state. The results from the Pauli gates are decoded and translated into applicable output data. A variational quantum circuits-based neural network plays various roles in QNN. The Adam optimizer updates the parameters based on various criteria such as the size of complexity-theoretic measurements, depth, accuracy, and definite features. The number of steps is necessary for solving the issue of in-depth measurement. Precision refers to the setup required to address a variety of challenges. A quantum neural network is composed of input, output, and L hidden layers. The L hidden layer consists of a quantum circuit of the quantum perceptron, which acts on an initial state of the input qubits and produces a mixed state for the output qubits. QNN can perform quantum computation for both the two-input and one-input qubit perceptron, which goes through the quantum-circuit construction with a quantum perceptron on 4 level qubits. The most comprehensive quantum perceptron implements any quantum channel on the input qubits. The precision of p(n) is represented by s (n), d(n), where size is denoted by s(n) and depth is denoted by d(n). Size refers to the number of qubits in the circuit, while depth refers to the longest sequence of gates from input to output. Size and depth are created from gates D and U of precision p(n). A reversible U gate is usually followed by the D gate to eliminate the localization problem. The accuracy of the circuits is denoted by Os(n) for evaluating the robustness of models against adversarial attacks.

## IV. EXPERIMENT AND RESULTS

In this section, we provide details of our adversarial attack experiments and results. We start by specifying the dataset used and the data processing techniques employed. In order to design effective adversarial attacks, we define the experimental settings, where we use accuracy [24], precision [25], recall [26], and F-score metrics to evaluate the robustness of models to attacks. Finally, we present the experimental results, which highlight the effectiveness of our adversarial attacks in compromising the performance of the targeted models.

### A. Dataset Specification

We utilized a Quantum Neural Network (QNN) to classify malware using the ClaMP dataset, which consists of two versions: ClaMP_raw and ClaMP_Integrated. ClaMP_raw was generated by aggregating instances from VirusShare, while ClaMP_Integrated contains both malware and benign instances gathered from Windows files. To extract features from the samples, we focused on the portable executable headers since they contain essential information required for the operating system to execute executable files. We collected various raw features from the PE headers, such as File Header (7 features), DOS header (19 features), and Optional header (29 features), using a rule-based method for both the malware and benign samples. Afterwards, significant features were obtained by utilizing the raw features such as entropy, compilation time, and section time. Furthermore, we extracted additional information about the PE file by expanding a collection of raw features from the file header. Subsequently, we chose three categories of features - raw, derived, and expanded - from the ClaMP_Integrated dataset. The dataset consisted of a total of 68 features, including 28 raw, 26 expanded, and 14 derived features [2].

### B. Data Preprocessing

We applied NN and QNN on ClaMP datasets to inspect the experimented method's comparative performance. We first considered the entire dataset. We separated 80 percent of the data from the 5210 instances to train the model in such a way that the software could supply it with the appropriate instance. It should be noted that within this 80 percent data, 60 percent was separated for training, 20 percent for validation, and 20 percent for testing. The perturbed data was generated by adding random noise to the clean data, altering its features. This mixture of clean and perturbed data was expected to confuse the model and potentially degrade its performance. The algorithm for adding perturbed data to the dataset is presented in Algorithm 1. The function *add_perturbation* takes two input parameters: the clean data and a noise scaling factor, epsilon. The noise is generated using a random number generator that creates an array with the same shape as the input data. This noise is then scaled by the epsilon factor, which determines the magnitude of the perturbations. Finally, the scaled noise is added to the original data, resulting in the perturbed data.

---
**Algorithm 1** Function to add random perturbations to input data.
**Function** `add_perturbation`(*data, epsilon*)
    noise ← np.random.randn(*data.shape)  scaled_noise ← epsilon * noise  perturbed_data ← data + scaled_noise
    **return** *perturbed_data*
**EndFunction**

---

### C. Experimental Settings

The current quantum simulator is unable to handle large dimensions as input, and our dataset has 108 dimensions, making it unsuitable for the simulator. As a result, we employed a dimension reduction technique called Principal Component Analysis (PCA) on this dataset. We applied PCA to the 108-feature vector of the CLaMP dataset to decrease the dimensionality. Due to the existing simulator's qubit number limitations, we chose the top 16 principal components. We first applied the classical neural network to the reduced dataset directly. Following that, we encoded the classical data as quantum circuits, which involves converting all feature values into qubit values for quantum computer processing.

We show the circuit produced for an arbitrary sample. These circuits were translated into TensorflowQuantum (TFQ) format. Subsequently, we designed a model circuit layer for the quantum neural network (QNN), consisting of a two-layer model with a data circuit size that matched the input. This model circuit was then wrapped in a TFQ-Keras model. We transformed the quantum data, fed it to the model, and employed a parametrized quantum layer to train the model circuit on the quantum data. During the training phase, we used an optimization function called hinge loss. We converted the labels to a range of -1 to 1. In the end, we trained the QNN for 100 epochs.

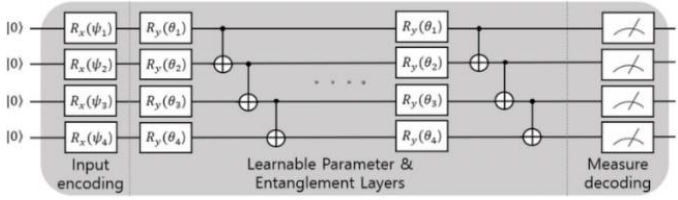

Fig. 1: Demonstrates the quantum neural network with the input parameter and Linear entanglement structure

*D. Experimental Results*

Our comparative analysis between the classical neural network (NN) model and the quantum neural network (QNN) model illustrates in table 1 and Table 2. We compared the performance of a classical neural network (NN) and a quantum neural network (QNN) using several metrics, including accuracy, precision, recall, and F1-score. Before the adversarial attack, we received NN: Accuracy - 0.54, Precision - 1.00, Recall - 0.57, F1-score - 0.65 and QNN: Accuracy - 0.57, Precision - 0.92, Recall - 0.57, F1-score - 0.65. The QNN model exhibited slightly higher accuracy (0.57) compared to the NN model (0.54). This indicates that the QNN model provided marginally better overall performance in terms of correctly predicting the class labels. Both models had similar recall and F1-score values.

After the adversarial attack, we recieved NN: Accuracy - 0.52, Precision - 0.27, Recall - 0.52, F1-score - 0.36 and QNN: Accuracy - 0.45, Precision - 0.47, Recall - 0.90, F1-score - 0.62. The performance of both models degraded, but the impact on each model was different. The NN model experienced a more significant drop in precision (from 1.00 to 0.27), while its accuracy and recall decreased only slightly. The F1-score for the NN model dropped to 0.36, indicating a substantial decrease in overall performance. In contrast, the QNN model experienced a reduction in accuracy (from 0.57 to 0.45) after the adversarial attack but demonstrated a remarkable increase in recall (from 0.57 to 0.90). This suggests that the QNN model was better at identifying true positives in the presence of adversarial data. The precision of the QNN model decreased from 0.92 to 0.47, and its F1-score dropped slightly to 0.62. The results indicate that the QNN model displayed greater resilience to adversarial attacks, maintaining a higher F1-score compared to the NN model. The QNN model's ability to achieve a higher recall value in the presence of adversarial data demonstrates its potential for robust performance in real-world applications where noisy or manipulated data might be present.

TABLE-1 : Results derived before adversarial attack

| Models | Accuracy | Precision | Recall | F1-Score |
|---|---|---|---|---|
| NN | 0.54 | 1.00 | 0.57 | 0.65 |
| QNN | 0.57 | 0.92 | 0.57 | 0.65 |

TABLE-2 : Results derived after adversarial attack

| Models | Accuracy | Precision | Recall | F1-Score |
|---|---|---|---|---|
| NN | 0.52 | 0.27 | 0.52 | 0.36 |
| QNN | 0.45 | 0.47 | 0.90 | 0.62 |

We evaluated the performance of the models showing plot of confusion matrix ( Figure 2 and 3), ROC Curve (Figure 4 and 5) and Precision Recall Curve ( Figure 6 and 7). The confusion matrix is a tabular representation that illustrates the distribution of predicted and true class labels, helping to identify the model's strengths and weaknesses in classification tasks. The rows in the confusion matrix represent the actual (true) class labels, while the columns represent the predicted class labels. The main diagonal of the matrix contains the counts of correctly classified instances, also known as true positives (TP) for each class. The off-diagonal elements indicate the misclassifications or errors made by the model [27]. From the confusion matrix in Figure 2 , we can observe the following insights: In the clean data scenario, the model performed well with no false positives and correctly classified 21 instances as class 0 and 4 instances as class 1. However, in the perturbed data scenario, the model still had no false positives, but it had a higher false negative rate, misclassifying 3 instances of class 1 as class 0, while correctly classifying only 1 instance of class 0. These results suggest that the adversarial attack caused a significant shift in the model's decision boundary, making it more susceptible to misclassifying some instances.

From the confusion matrix in Figure 3, we can observe the following insights: The first confusion matrix represents the distribution of predicted and true class labels for a binary classification problem using clean data, prior to any adversarial attack. The matrix shows that there were 160 instances of class 1 and 210 instances of class 2. The model correctly classified all instances of class 0, resulting in 0 false positives. However, the model misclassified all 210 instances of class 1 as class 2, resulting in 210 false negatives.

The second confusion matrix represents the distribution of predicted and true class labels for a binary classification problem using perturbed data after the adversarial attack. The matrix shows that there were 55 instances of class 1 and 39 instances of class 2. The model correctly classified all instances of class 0, resulting in 0 false positives. However, the model misclassified 39 instances of class 1 as class 2, resulting in 39 false negatives.

The comparison of the two confusion matrices shows that the adversarial attack had a significant impact on the model's performance. In the clean data scenario, the model was unable to correctly classify any instances of class 1, while in the perturbed data scenario, the model was able to correctly classify some instances of class 1 but still had a high false-

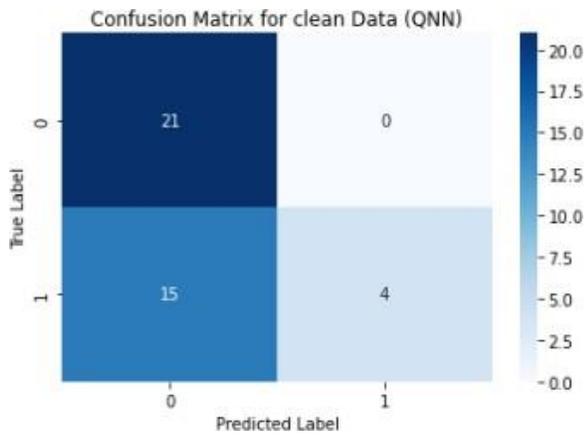

((a)) Confusion matrix for Clean Data (QNN)

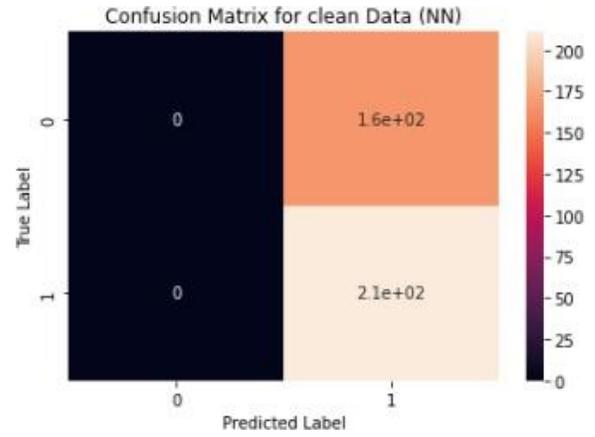

((a)) Confusion matrix for Clean Data (NN)

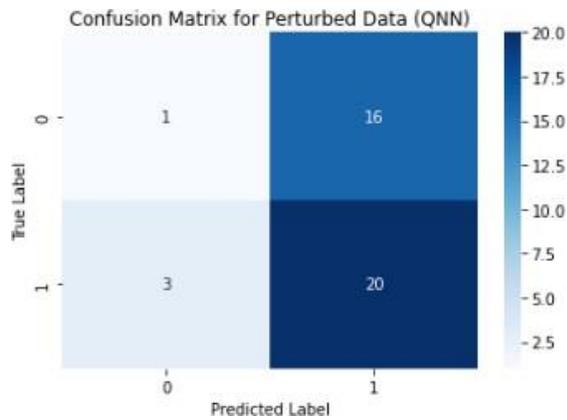

((b)) Confusion matrix for Perturbed Data (QNN)

Fig. 2: Comparison of Confusion Matrix between Clean Data and Perturbed Data of QNN Model.

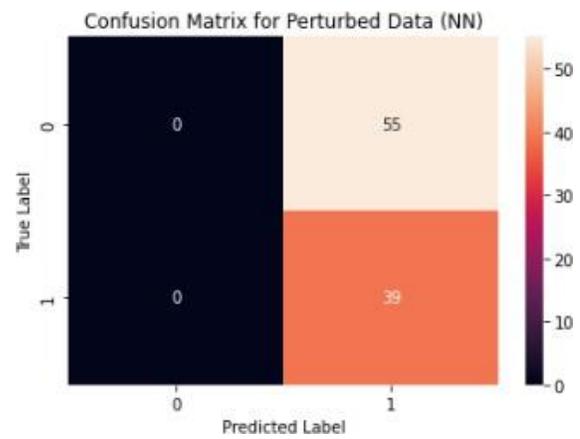

((b)) Confusion matrix for Perturbed Data (NN)

Fig. 3: Comparison of Confusion Matrix between Clean Data and Perturbed Data of NN Model.

negative rate. These results suggest that the model needs further training or fine-tuning to improve its performance on adversarial data.

The ROC (Receiver Operating Characteristic) curve is a graphical representation that measures the performance of a binary classifier as the discrimination threshold varies. The ROC curve plots the true positive rate (TPR) against the false positive rate (FPR) at various threshold settings, where the TPR is the proportion of positive instances that are correctly classified, and the FPR is the proportion of negative instances that are incorrectly classified as positive.

The ROC curve is helpful for measuring the performance of a binary classifier because it provides a way to visualize and compare the trade-off between the sensitivity (TPR) and specificity (1-FPR) of the classifier at different threshold settings. A perfect classifier would have a TPR of 1 and an FPR of 0, resulting in a point at the top left corner of the ROC curve. In contrast, a random classifier would have a diagonal line from the bottom left to the top right of the ROC curve, where the area under the curve (AUC) is 0.5, indicating that the classifier is no better than random guessing.

A higher AUC indicates a better performance of the classifier, where an AUC of 1 represents a perfect classifier, while an AUC of 0.5 indicates a random classifier. The ROC curve can also help to determine the optimal threshold setting for the classifier, depending on the desired balance between the TPR and FPR [28].

From the ROC curve in Figure 4, we can observe the following insights: The first ROC curve represents the performance of the QNN model on clean data before any adversarial attack. The AUC value of 0.61 indicates that the model has a moderate performance in correctly classifying the positive and negative instances. The curve starts at the bottom left corner (0,0), indicating that the model correctly classified all negative instances, but misclassified some positive instances as negative. As the threshold increases, the TPR increases at a faster rate than the FPR, resulting in an upward curve. However, the curve is not very steep, indicating that the model's performance is not very sensitive to changes in the threshold.

The second ROC curve represents the performance of the QNN model on perturbed data after the adversarial attack. The AUC value of 0.46 indicates that the model's performance has degraded significantly after the attack. The curve starts at the bottom left corner (0,0), indicating that the model correctly classified all negative instances, but it misclassified a significant number of positive instances as negative. As the threshold increases, the TPR increases at a much slower rate than the FPR, resulting in a nearly linear curve. The curve has a slight upward increment towards the top right corner (1,1), indicating that the model is able to correctly classify some positive instances, but it is not very sensitive to changes in the threshold.

The comparison of the two ROC curves highlights the impact of the adversarial attack on the performance of the QNN model. The AUC value decreased significantly from 0.61 to 0.46, indicating that the model's ability to correctly classify positive and negative instances has deteriorated. The second curve also shows a higher FPR at lower TPR, indicating that the model is more prone to false positives and less effective in detecting positive instances. These results suggest that the QNN model needs further improvement in order to be more robust against adversarial attacks. Additional evaluation metrics such as precision, recall, and F1 score should be computed to provide a more comprehensive evaluation of the model's performance.

From the ROC curve in Figure 5, we can observe the following insights: The two ROC curves represent the performance of the NN model on clean and perturbed data before and after the adversarial attack, respectively. Both ROC curves have an AUC value of 0.50, which indicates that the model's performance is equivalent to random guessing. The curves are nearly linear, starting at the bottom left corner (0,0) and ending at the top right corner (1,1). This indicates that the model has an equal probability of correctly classifying positive and negative instances and is not effective in distinguishing between them.

These results suggest that the NN model is not suitable for the binary classification task, and the adversarial attack did not significantly impact the model's performance.

The Precision-Recall curve is a graphical representation of the performance of a binary classifier at different thresholds. It plots the precision against the recall at various threshold settings, where precision is the proportion of true positive instances out of all instances predicted as positive, and recall is the proportion of true positive instances that are correctly classified. The Precision-Recall curve is important for performance measurement in ML models because it provides a more informative evaluation of a binary classifier's performance compared to other metrics such as accuracy or F1 score, especially in imbalanced datasets where the number of positive instances is significantly smaller than the number of negative instances. The Precision-Recall curve can help to determine the optimal threshold setting for the classifier, depending on the desired trade-off between precision and recall. A high

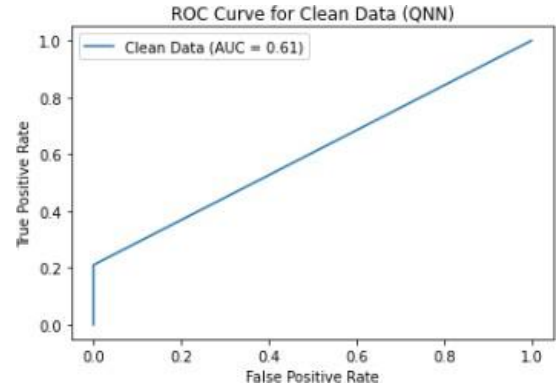

((a)) ROC Curve for Clean Data (QNN)

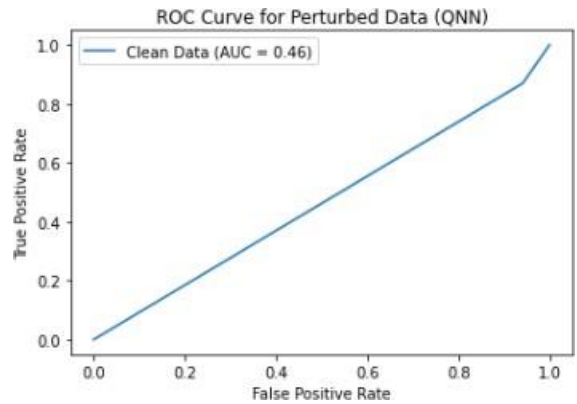

((b)) ROC Curve for Perturbed Data (QNN)

Fig. 4: Comparison of ROC Curve between Clean Data and Perturbed Data of QNN Model.

precision means that the classifier correctly identifies most positive instances with few false positives, while a high recall means that the classifier correctly identifies most positive instances out of all positive instances [29]. From the Precision Recall curve in Figure 6, we can observe the following insights:

The two precision-recall curves represent the performance of the QNN model on clean and perturbed data before and after the adversarial attack, respectively. The first curve, for clean data, has an AUC value of 0.79, which indicates that the model's performance in correctly identifying positive instances is relatively good. The curve starts at the top left corner (1.0,0.0), indicating that the model correctly identified all positive instances, but it had some false positives. As the recall decreases, the precision decreases gradually, resulting in a curve that is mostly flat but with a slight decline towards the bottom right corner (0.0,1.0).

The second curve, for perturbed data after the adversarial attack, has an AUC value of 0.75, which indicates that the model's performance has slightly degraded. The curve starts at the top left corner (1.0,0.85), indicating that the model correctly identified most positive instances, but it had some

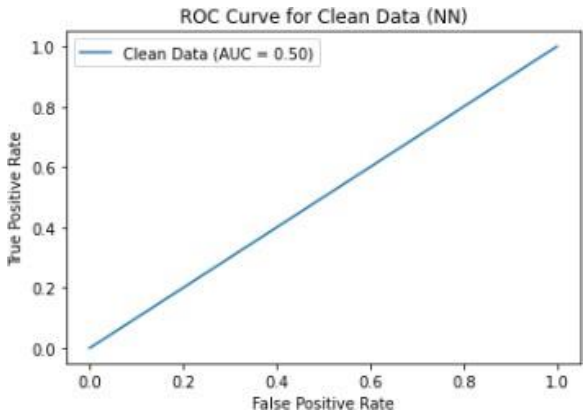

((a)) ROC Curve for Clean Data (NN)

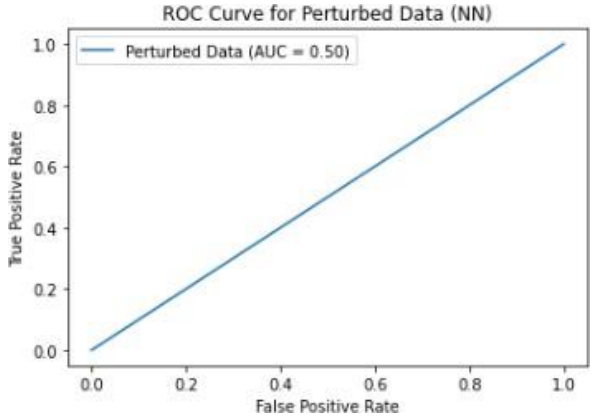

((b)) ROC Curve for Perturbed Data (NN)

Fig. 5: Comparison of ROC Curve between Clean Data and Perturbed Data of NN Model.

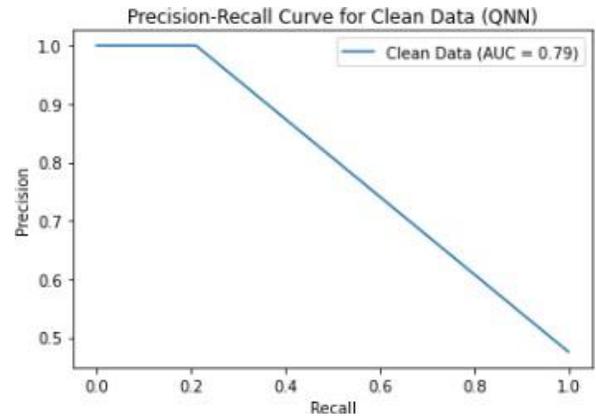

((a)) Precision Recall Curve for Clean Data (QNN)

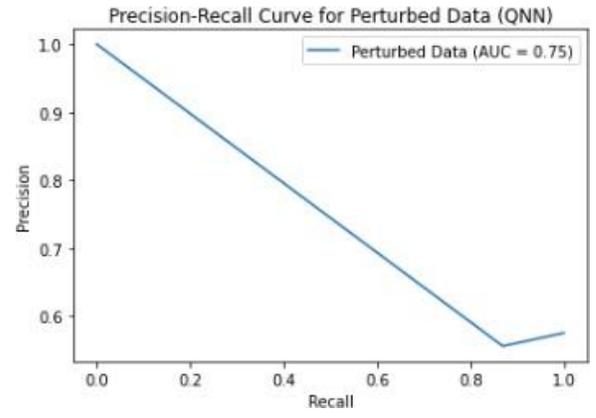

((b)) Precision Recall Curve for Perturbed Data (QNN)

Fig. 6: Comparison of Precision Recall Curve between Clean Data and Perturbed Data of QNN Model.

false positives. As the precision decreases, the recall increases, resulting in a curve that increases gradually towards the bottom right corner (1.0,0.6).

These results suggest that the adversarial attack had some impact on the QNN model's performance, but the model is still able to effectively identify positive instances, albeit with a higher false-positive rate.

From the Precision Recall curve in Figure 7, we can observe the following insights:

The two precision-recall curves represent the performance of the NN model on clean and perturbed data before and after the adversarial attack, respectively. The first curve, for clean data, has an AUC value of 0.78, which indicates that the model's performance in correctly identifying positive instances is relatively good. The curve starts at the top left corner (1.0,0.0), indicating that the model correctly identified all positive instances, but it had some false positives. As the precision decreases, the recall decreases gradually, resulting in a curve that is mostly linear but with a slight decline towards the bottom right corner (0.0,1.0).

The second curve, for perturbed data after the adversarial attack, has an AUC value of 0.71, which indicates that the model's performance has slightly degraded. The curve starts at the top left corner (1.0,0.0), indicating that the model correctly identified all positive instances, but it had some false positives. As the precision decreases, the recall increases, resulting in a curve that decreases linearly towards the bottom right corner (1.0,0.6).

These results suggest that the adversarial attack had some impact on the NN model's performance, but the model is still able to effectively identify positive instances, albeit with a higher false-positive rate

## V. DISCUSSION

In this study, we have conducted a comparative analysis of the performance of classical neural network (NN) and quantum neural network (QNN) models for a binary classification task. Our goal was to evaluate their resilience against adversarial attacks and to understand their potential for real-world applications, where the presence of noisy or manipulated data is likely. We compared the performance of both models using various metrics, including accuracy, precision, recall, and F1-score, on clean and perturbed datasets. Our results indicate

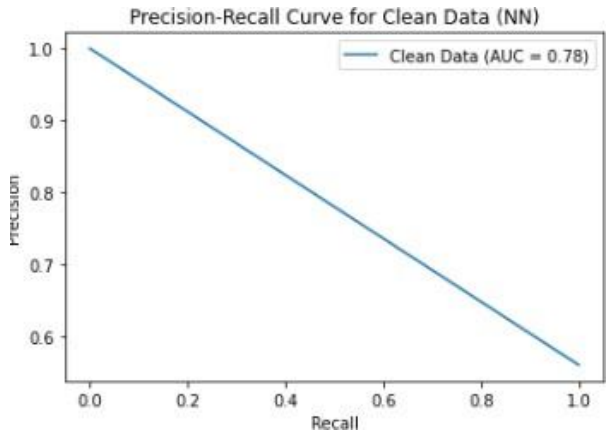

((a)) Precision Recall Curve for Clean Data (NN)

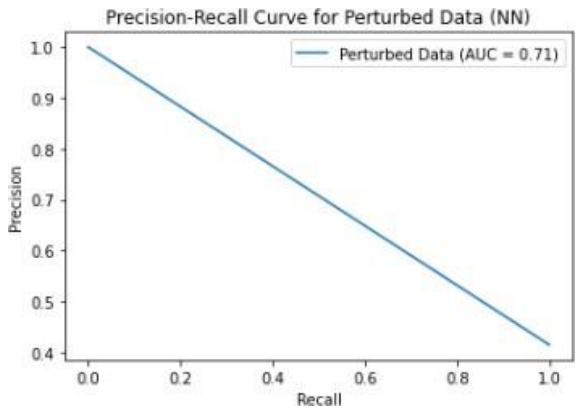

((b)) Precision Recall Curve for Perturbed Data (NN)

Fig. 7: Comparison of Precision Recall Curve between Clean Data and Perturbed Data of NN Model.

that the QNN model exhibited slightly higher accuracy in the clean data scenario compared to the NN model, with similar recall and F1-score values. This suggests that the QNN model provided marginally better overall performance in terms of correctly predicting the class labels. However, after the adversarial attack, the performance of both models degraded, but the impact on each model was different. The NN model experienced a more significant drop in precision, while the QNN model demonstrated a remarkable increase in recall. This indicates that the QNN model was better at identifying true positives in the presence of adversarial data, which demonstrates its potential for robust performance in real-world applications. The comparison of confusion matrices, ROC curves, and precision-recall curves for both models before and after the adversarial attack further supports our findings. The confusion matrices reveal that the QNN model maintained a higher true positive rate and lower false-negative rate compared to the NN model after the adversarial attack, indicating its greater resilience. The ROC curves show that the QNN model's performance in terms of distinguishing between positive and negative instances has deteriorated after the attack, but it still outperformed the NN model. The precision-recall curves indicate that the QNN model's ability to identify positive instances effectively

## VI. CONCLUSION

This paper presents a novel comparative analysis of the vulnerability of machine learning (ML) and quantum machine learning (QML) models, specifically conventional neural networks (NN) and quantum neural networks (QNN), to adversarial attacks using a malware dataset from the software supply chain domain. The study is among the first to focus on software supply chain vulnerabilities using quantum machine learning, and employs Pennylane, a cutting-edge quantum computing platform that allows seamless integration with various QML tools. Our contributions include the adoption of both quantum and conventional machine learning approaches to conduct experiments on a software supply chain attack dataset, and an assessment of the performance of NN and QNN models under adversarial attack scenarios, enabling a comparison of their vulnerability and robustness. The findings reveal that both ML and QML models are susceptible to adversarial attacks, but QNNs demonstrate higher resilience in certain aspects. The outcomes of this study contribute to a better understanding of the challenges and opportunities in the deployment of ML and QML models in security-sensitive domains, and provide a foundation for further research aimed at enhancing the resilience of these models to adversarial attacks. In future we willconduct a more extensive range of experiments to better understand the performance and robustness of both conventional and quantum neural networks under various adversarial conditions.

ACKNOWLEDGEMENT

The work is supported by the National Science Foundation under NSF Award #2209638, and #2100115, Any opinions, findings, recommendations, expressed in this material are those of the authors and do not necessarily reflect. the views of the National Science Foundation.